\documentclass[10pt,twocolumn,letterpaper]{article}

\usepackage{cvpr}              % To produce the CAMERA-READY version
\usepackage[accsupp]{axessibility}  % Improves PDF readability for those with disabilities.

\usepackage{graphicx}
\usepackage{amsmath}
\usepackage{amssymb}
\usepackage{booktabs}
\usepackage{xcolor}
\usepackage{multirow}
\usepackage{makecell}
\usepackage{wrapfig}
\usepackage{colortbl}
\usepackage{tikz}

\definecolor{darkgreen}{HTML}{11823b}
\definecolor{green}{HTML}{48bf53}
\definecolor{lightgreen}{HTML}{91f086}

\definecolor{lightred}{HTML}{F6BDC0}

\definecolor{gold}{HTML}{FBF2D2}
\definecolor{silver}{HTML}{DDDDDD}
\definecolor{bronze}{HTML}{EED2B8}

\definecolor{goldD}{HTML}{D9AE13}
\definecolor{silverD}{HTML}{909090}
\definecolor{bronzeD}{HTML}{9A5F26}

\definecolor{MS}{HTML}{b85450}
\definecolor{optical}{HTML}{6c8ebf}
\definecolor{student}{HTML}{1c8d39}
\definecolor{augs}{HTML}{9673a6}

\definecolor{falseNeg}{HTML}{1E88E5}
\definecolor{falsePos}{HTML}{E4AD06}

\newcommand{\medal}[3]{\tikz[baseline=(char.base)]{\node[rounded corners=2pt,fill=#1,draw=#2,inner sep=1.5pt](char){#3};}}

\newcommand{\bm}[2]{
    \ifcase#1\or% case 1
      {\medal{gold}{goldD}{\textbf{#2}}}
    \or % case 2
      {\medal{silver}{silverD}{#2}}
    \or % case 3
      {\medal{bronze}{bronzeD}{#2}}
    \else % default case
      #2
    \fi\ignorespaces
}

\definecolor{cvprblue}{rgb}{0.21,0.49,0.74}
\usepackage[pagebackref,breaklinks,colorlinks,allcolors=cvprblue]{hyperref}

\def\paperID{*****} % *** Enter the Paper ID here
\def\confName{CVPR}
\def\confYear{2026}

\title{Brewing Stronger Features: \\ Dual-Teacher Distillation for Multispectral Earth Observation}

\author{Filip Wolf
\and Blaž Rolih
\and Luka Čehovin Zajc
\and University of Ljubljana, Faculty of Computer and Information Science, Slovenia\\
{\tt\small filip.wolf@fri.uni-lj.si}
}

\begin{document}
\maketitle
\begin{abstract}
Foundation models are transforming Earth Observation (EO), yet the diversity of EO sensors and modalities makes a single universal model unrealistic. Multiple specialized EO foundation models (EOFMs) will likely coexist, making efficient knowledge transfer across modalities essential. Most existing EO pretraining relies on masked image modeling, which emphasizes local reconstruction but provides limited control over global semantic structure. To address this, we propose a dual-teacher contrastive distillation framework for multispectral imagery that aligns the student’s pretraining objective with the contrastive self-distillation paradigm of modern optical vision foundation models (VFMs). Our approach combines a multispectral teacher with an optical VFM teacher, enabling coherent cross-modal representation learning. Experiments across diverse optical and multispectral benchmarks show that our model adapts to multispectral data without compromising performance on optical-only inputs, achieving state-of-the-art results in both settings, with average improvements of 3.64 percentage points in semantic segmentation, 1.2 in change detection, and 1.31 in classification. This demonstrates that contrastive distillation provides a principled and efficient approach to scalable representation learning across heterogeneous EO data sources. Project page: \textcolor{magenta}{https://wolfilip.github.io/DEO/}.
\end{abstract}    
\section{Introduction}
\label{sec:intro}

Foundation models (FMs) have recently emerged as a powerful paradigm in Earth Observation (EO), demonstrating strong transferability across diverse downstream tasks~\cite{wang2025towards,danish2025terrafm,fayad2025dunia}. They leverage large volumes of unlabeled data~\cite{waldmann2025panopticon,zhang2025skysense,wang2025towards}, reduce reliance on scarce or inconsistent labels~\cite{do2025robsense,jia2025can,tseng2025galileo,brown2025alphaearth}, and enable flexible task adaptation~\cite{rolf2024position,xiao2025foundation,bodnar2025foundation}. These qualities are particularly valuable in EO, where data collection is abundant but high-quality annotations are limited~\cite{li2023cost}.

\begin{figure}[!t]
    \centering
    \includegraphics[width=\linewidth]{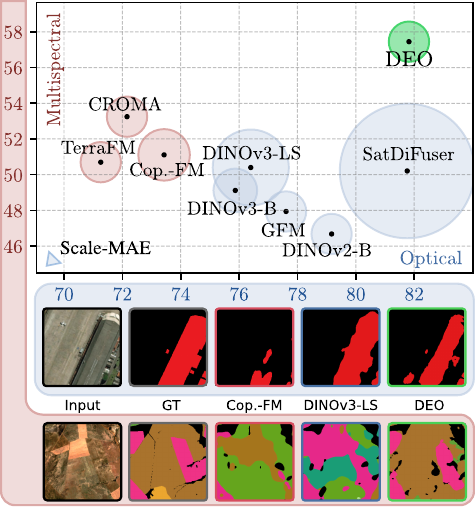}
    \caption{\textbf{DEO}, our proposed dual-teacher pretraining approach, results in a model that achieves state-of-the-art performance in \textcolor{MS}{multispectral} EO tasks while maintaining performance on \textcolor{optical}{optical} EO tasks. On top, we demonstrate our performance in optical and multispectral semantic segmentation, visualizing model size using colored circles. Below, the first row of images shows qualitative results for the optical SpaceNetv1~\cite{van2018spacenet} dataset, while the second row shows results for the multispectral m-SA-crop-type dataset~\cite{lacoste2023geo}.}
    \label{fig:intro}
    \vspace{-0.75em}
\end{figure}

However, developing a single, universal EO foundation model (EOFM) remains challenging~\cite{xiong2024one}. EO data vary widely in spatial resolution, spectral characteristics, and acquisition conditions, while geographic and seasonal variations further introduce domain shifts~\cite{xiong2024one,astruc2025anysat,reed2023scale,xiao2025foundation,brown2025alphaearth}, rendering the training of a single model that effectively captures the intricacies of various EO data difficult. Instead, progress increasingly depends on \emph{efficient knowledge transfer} between models~\cite{barseghyan2025less}, and knowledge distillation provides a practical mechanism for achieving this.

EOFMs for multispectral (MS) satellite imagery are a strong target for improvement by distillation. They contain the optical (RGB) channels that can be well-represented by modern general-purpose vision foundation models (VFMs)~\cite{waldmann2025panopticon,simeoni2025dinov3}, which are prime candidates for knowledge extraction. On the other hand, EOFMs can also additionally encode rich spectral information from MS data, which is required for many EO applications. Since training a new EOFM from scratch on MS data is computationally expensive~\cite{oquab2024dinov,simeoni2025dinov3}, leveraging existing VFMs becomes an appealing alternative. However, knowledge transfer must be done correctly to maximize its effectiveness.

In the broader computer vision community, modern large VFMs are typically trained using contrastive and self-distillation objectives~\cite{caron2021emerging,oquab2024dinov,simeoni2025dinov3} that explicitly shape global semantics and produce latent feature spaces well-suited for downstream transfer. In contrast, much of EO pretraining still relies on masked image modeling (MIM)~\cite{cong2022satmae,wang2025towards,mendieta2023towards}, which emphasizes local reconstruction~\cite{he2022masked, xie2022simmim} and imposes weaker constraints on global semantic structure. Consequently, contrastive and self-distillation approaches remain comparatively underexplored in EO~\cite{danish2025terrafm,waldmann2025panopticon,zhang2025skysense}. Motivated by the observation, we propose a \textbf{contrastive dual-teacher} distillation framework for MS representation learning. Our method pairs \textbf{(i)} a contrastive self-distillation multispectral teacher that structures the MS feature space and \textbf{(ii)} an optical VFM teacher that provides high-level semantic priors learned at a global scale. Because the student and the optical teacher share compatible pretraining objectives, the student's latent space more readily matches that of the VFM compared to pairing distillation with an MIM objective. This compatibility yields more coherent cross-modal transfer and better downstream performance, shown in \Cref{fig:intro}. Our contributions are as follows:

\begin{itemize}
    \item We introduce a \textit{dual-teacher pretraining strategy} that unifies a contrastive self-distillation multispectral teacher with distillation from an optical teacher, combining global representation learning with transfer of semantic priors.
    \item We demonstrate that matching the student's pretraining objective with that of a VFM teacher (e.g., DINOv3) enables a more effective and data-efficient transfer of optical priors to a multispectral student.
\end{itemize}

Our model, \textbf{DEO} (\textit{\textbf{D}istillation for \textbf{E}arth \textbf{O}bservation}), achieves state-of-the-art performance across optical and multispectral downstream tasks, with an average improvement of $3.64$ percentage points in semantic segmentation, $1.2$ points in change detection, and $1.31$ points in classification. This highlights distillation-centric training as a key strategy for building a sustainable and interoperable landscape of EO foundation models.
\section{Related Work}
\label{sec:formatting}

\noindent\textbf{Contrastive learning and distillation.} Contrastive Learning (CL) is a widely used technique for self-supervised learning responsible for many breakthroughs in the field~\cite{chen2020simple,grill2020bootstrap,he2020momentum,caron2021emerging,bardes2022vicreg,zbontar2021barlow}. In contrast to MIM-based methods such as Masked Autoencoders (MAE)~\cite{xie2022simmim,he2022masked,balestriero2024learning}, CL leads to strong semantic representations invariant to distribution shifts. While sensitive to the choice of data augmentation and prone to dimensional collapse, recent advances have aimed to mitigate these drawbacks~\cite{wu2025simplifying}.

As large pretrained foundation models rose in prominence, so did the concept of distillation. It is primarily used to transfer knowledge between models, with usage being divided into compressing large pretrained models into smaller ones~\cite{oquab2024dinov,simeoni2025dinov3,wu2025simplifying,sariyildiz2025dune,oord2018representation,heinrich2025radiov2,ranzinger2024radio} or improving newer models~\cite{heinrich2025radiov2,ranzinger2024radio,mendieta2023towards,wang2025towards,han2024bridging,liu2022multispectral}. Recently, the closely-related representation learning technique of self-distillation was used to train advanced VFMs~\cite{caron2021emerging,zhou2021ibot,oquab2024dinov,simeoni2025dinov3}.

\noindent\textbf{Vision Foundation Models.} General-purpose VFMs have demonstrated strong performance across many downstream tasks with minimal fine-tuning, even without task-specific training~\cite{mendieta2023towards,wang2024multi,gong2025crossearth,waldmann2025panopticon}. This is largely due to a key commonality between these models: extensive pretraining on large image corpora and careful data curation~\cite{radford2021learning,kirillov2023segment,oquab2024dinov,simeoni2025dinov3}. Recent large VFMs, such as DINOv2~\cite{oquab2024dinov} and DINOv3~\cite{simeoni2025dinov3}, were trained on more than $100$ million curated images and have demonstrated a capacity to consistently improve results across various domains~\cite{baharoon2023evaluating,wanyan2024extending}, with DINOv3 showing a strong focus on the EO domain. The development of VFMs pushes boundaries not only in terms of methodology but also in protocols for input data collection and preparation. Effectively utilizing the strong general representations present in VFMs is a great asset for many specialized domains, including EO.

\begin{figure*}[t]
    \centering
    \includegraphics[width=\textwidth]{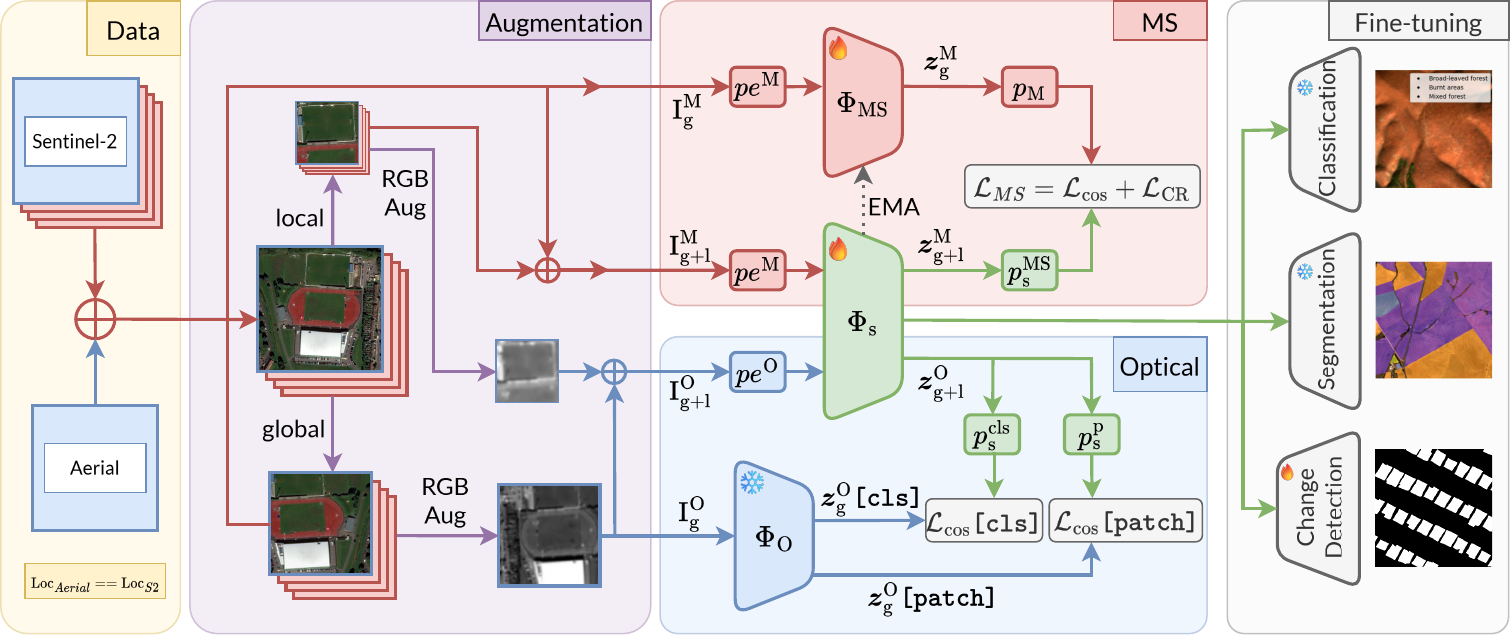}
    \caption{\textbf{Overview of our dual-teacher pretraining approach.} The pretraining dataset utilizes a standard fMoW Sentinel-2 dataset, augmented with high-resolution aerial images where possible. Random crops and other augmentation operations are performed on sampled images. Full multispectral and optical channel subsets are used for their corresponding distillation branches. The multispectral branch is a contrastive learning setup where the teacher is updated using EMA. In the optical branch, distillation is done using a frozen VFM teacher. The resulting model can then be used in various downstream tasks.}
    \label{fig:method}
    \vspace{-0.75em}
\end{figure*}

\noindent\textbf{Earth Observation Foundation Models.} Most EOFM proposed in recent years were pretrained using MIM-based techniques~\cite{reed2023scale,cong2022satmae,li2024masked,wang2025towards,mendieta2023towards,xiong2024neural,astruc2024omnisat} and an ever-increasing amount of unlabeled EO data. Less common approaches explore alternative representation learning techniques, such as CL~\cite{tseng2025galileo,danish2025terrafm,fuller2023croma,do2025robsense,fayad2025dunia,waldmann2025panopticon}, JEPA~\cite{astruc2025anysat}, and diffusion~\cite{khanna2024diffusionsat,jia2025can}. Pioneering methods such as Scale-MAE~\cite{reed2023scale}, while notable for establishing the field, were limited to optical data and simple MAE-style pretraining. CROMA~\cite{fuller2023croma} was among the first to utilize CL to achieve strong results by combining modalities, while recently, TerraFM~\cite{danish2025terrafm} and SatDiFuser~\cite{jia2025can} have used modern representation learning techniques on diverse input data to achieve strong results. 

Due to the perceived domain shift, VFM distillation has only recently been explored in EO~\cite{mendieta2023towards,han2024bridging}. The recently proposed Copernicus-FM~\cite{wang2025towards} combines MIM-style pretraining with DINOv2-based distillation. However, its reconstruction-driven objective is not fully aligned with the contrastive and distillation losses used to train modern VFMs~\cite{oquab2024dinov,simeoni2025dinov3}, resulting in a weaker global semantic structure of features. In contrast, we integrate VFM distillation into a contrastive pretraining pipeline, ensuring objective-level consistency with the teacher.
\section{Methodology}
\label{sec:method}

We propose a pretraining approach that excels at diverse downstream tasks when multispectral data is available, without compromising performance on tasks that use only the optical bands (i.e., RGB) of an image. As shown in Figure~\ref{fig:method}, our approach uses a contrastive self-distillation student-teacher framework with \textit{two} teachers. One teacher specializes in teaching students optical representations, while the other specializes in teaching students robust multispectral representations. Together, they train a single student network to excel in both input modalities. To aid in clarity, we use different colored notation for different parts of the network: a \textcolor{MS}{red} color for the \textcolor{MS}{multispectral} teacher, \textcolor{optical}{blue} for the \textcolor{optical}{optical} teacher, and \textcolor{student}{green} for the \textcolor{student}{student}. In the following sections, we provide an introduction to contrastive self-distillation and describe each component of the network.

\subsection{Contrastive self-distillation}
\label{subsec:cl}

The core part of our method (\textcolor{MS}{red} section in \Cref{fig:method}) is based on DINO~\cite{caron2021emerging}, a self-supervised pretraining method designed for learning global data representations. DINO is a \textit{contrastive} method: it aligns representations generated from different \textit{views} (i.e., various crops) of the same input image and ensures their similarity. It is also \textit{self-distilling}, as the teacher's weights are an \textit{exponential moving average} (EMA) of the student's weights, which are updated in turn using backpropagation. This mechanism provides effective global representation learning, while also preventing the teacher's and student's weights from diverging~\cite{chen2021exploring,wu2025simplifying,grill2020bootstrap,he2020momentum}.

We enforce similarity between representations of different image views using a \textit{compression loss} such as cosine similarity. This, however, can lead to \textit{representation collapse} on its own, where both the teacher and student predict constant vectors (e.g., zero vectors)~\cite{grill2020bootstrap}, since no constraints are placed on the structure of the embedding space. To prevent this from happening, we additionally use an \textit{expansion loss} in the form of a \textit{coding rate regularizer}~\cite{wu2025simplifying}. The main idea is to maintain diversity in the representation space by penalizing a low-rank covariance matrix of the features, forcing the model to spread information across all feature dimensions rather than collapsing to trivial vectors. Concretely, we do this by forcing the log-determinant of the diagonal between the covariance matrix of the combined student and teacher output features $\boldsymbol{z}$ and the unit vector to be as large as possible, i.e., by minimizing $\mathcal{L}_\text{CR}:= -\log\det \left( \boldsymbol{I} + \operatorname{Cov}[\boldsymbol{z}] \right)$. Further methodology details are available in the Supplementary.

\subsection{Crafting diverse inputs}
\label{subsec:input}

We generate diverse augmented views of input images (\textcolor{augs}{purple} section in \Cref{fig:method}) to learn a stronger feature space~\cite{chen2020simple,zbontar2021barlow}. We use 10-channel Sentinel-2 imagery \textcolor{MS}{$\text{I}^\text{M}$} as input, where the first three channels represent the optical (RGB) part of an image, i.e. $\textcolor{optical}{\text{I}^\text{O}}\subset\textcolor{MS}{\text{I}^{\text{M}}}$ (we provide more pretraining data details in \Cref{subsec:pretraining}). For each $\textcolor{MS}{\text{I}^{\text{M}}}$, the following augmentation pipeline is applied. We first lightly augment $\textcolor{MS}{\text{I}^{\text{M}}}$ with channel-agnostic augmentations like flipping, denoted as $A_\text{M}$. We then additionally augment $\textcolor{optical}{\text{I}^\text{O}}$ with heavy augmentations $A_\text{O}$, such as color jitter, Gaussian blur, and solarization, to ensure robustness to input perturbations.

From this augmented image, we create $n$ larger \textit{global} views $\textcolor{MS}{\text{I}^\text{M}_\text{g}}$ and $m$ smaller \textit{local} views $\textcolor{MS}{\text{I}^\text{M}_\text{l}}$ using cropping and resizing. To teach MS representations, we concatenate $\textcolor{MS}{\text{I}^\text{M}_\text{g}}$ and $\textcolor{MS}{\text{I}^\text{M}_\text{l}}$ into $\textcolor{MS}{\text{I}^\text{M}_{\text{g}+\text{l}}}$ for the student, while the MS teacher obtains only the global views $\textcolor{MS}{\text{I}^\text{M}_\text{g}}$. For teaching optical representations, we use the optical images $\textcolor{optical}{\text{I}^\text{O}_{\text{g}+\text{l}}}$ for the student, while the teacher again receives only the global part $\textcolor{optical}{\text{I}^\text{O}_\text{g}}$. Formally:
\begin{equation}
\textcolor{MS}{\text{I}^\text{M}_{\text{g}+\text{l}}} = \textcolor{MS}{\text{I}^\text{M}_\text{g}} +\textcolor{MS}{\text{I}^\text{M}_\text{l}}, \textcolor{optical}{\text{I}^\text{O}_{\text{g}+\text{l}}} = \textcolor{optical}{\text{I}^\text{O}_\text{g}} +\textcolor{optical}{\text{I}^\text{O}_\text{l}}, \textcolor{optical}{\text{I}^\text{O}_{\text{g}+\text{l}}}\subset\textcolor{MS}{\text{I}^\text{M}_{\text{g}+\text{l}}},
\end{equation}
where
\begin{equation}
\begin{gathered}
\textcolor{MS}{\text{I}^\text{M}_\text{g}}=g(A_\text{M}(\textcolor{MS}{\text{I}^\text{M}})),
\textcolor{MS}{\text{I}^\text{M}_\text{l}}=l(A_\text{M}(\textcolor{MS}{\text{I}^\text{M}})),\\
\textcolor{optical}{\text{I}^\text{O}_\text{g}}=g(A_\text{O}(\textcolor{optical}{\text{I}^\text{O}}\subset A_\text{M}(\textcolor{MS}{\text{I}^\text{M}}))),\\
\textcolor{optical}{\text{I}^\text{O}_\text{l}}=l(A_\text{O}(\textcolor{optical}{\text{I}^\text{O}}\subset A_\text{M}(\textcolor{MS}{\text{I}^\text{M}}))).
\end{gathered}
\end{equation}
Here, $g$ is the function used for creating global views, and $l$ is for local views. All augmentations are performed randomly for each input image.

\subsection{Learning multispectral representations}
\label{subsec:ms}

We utilize the method described in \Cref{subsec:cl} to learn strong global multispectral representations, shown in \textcolor{MS}{red} in \Cref{fig:method}. We start by passing $\textcolor{MS}{\text{I}^\text{M}_\text{g}}$ to the multispectral teacher encoder $\textcolor{MS}{\Phi_\text{MS}}$ and $\textcolor{MS}{\text{I}^\text{M}_{\text{g}+\text{l}}}$ to the student encoder $\textcolor{student}{\Phi_\text{s}}$ (notice that $\textcolor{student}{\Phi_\text{s}}$ receives both local and global image views), employing a 10-channel patch embedding layer $\textcolor{MS}{pe^\text{M}}$ before the encoders. $\textcolor{MS}{\Phi_\text{MS}}$ and $\textcolor{student}{\Phi_\text{s}}$ produce features $\textcolor{MS}{\Phi_\text{MS}}(\textcolor{MS}{\text{I}^\text{M}_\text{g}}) = \textcolor{MS}{\boldsymbol{z}^\text{M}_\text{g}}$ and $\textcolor{student}{\Phi_\text{s}}(\textcolor{MS}{\text{I}^\text{M}_{\text{g}+\text{l}}}) = \textcolor{student}{\boldsymbol{z}^\text{M}_{\text{g}+\text{l}}}$, which are then projected into a common feature space via projection heads $\textcolor{MS}{p_\text{M}}$ for $\textcolor{MS}{\Phi_\text{MS}}$ and $\textcolor{student}{p_\text{s}^\text{MS}}$ for $\textcolor{student}{\Phi_\text{s}}$. Adding small projection heads improves performance and training stability compared to computing loss directly on backbone outputs~\cite{caron2021emerging}. We can then formulate the loss function for learning MS features,  as
\begin{equation}
\begin{aligned}
\textcolor{MS}{\mathcal{L}_{MS}} 
& = \mathcal{L}_\text{cos}(\textcolor{MS}{p_\text{M}}(\textcolor{MS}{\boldsymbol{z}^\text{M}_\text{g}}), \textcolor{student}{p_\text{s}^\text{MS}}(\textcolor{student}{\boldsymbol{z}^\text{M}_{\text{g}+\text{l}}}))\\
& - \gamma\mathcal{L}_\text{CR}(\textcolor{MS}{p_\text{M}}(\textcolor{MS}{\boldsymbol{z}^\text{M}_\text{g}}), \textcolor{student}{p_\text{s}^\text{MS}}(\textcolor{student}{\boldsymbol{z}^\text{M}_{\text{g}+\text{l}}})),
\end{aligned}
\label{eq:ms}
\end{equation}
where $\mathcal{L}_\text{cos}$ represents \textit{cosine similarity}, $\mathcal{L}_\text{CR}$ is the \textit{coding rate regularizer} described in \Cref{subsec:cl}, and $\gamma$ is a weighting coefficient.

\begin{figure}[t]
  \centering
   \includegraphics[width=\linewidth]{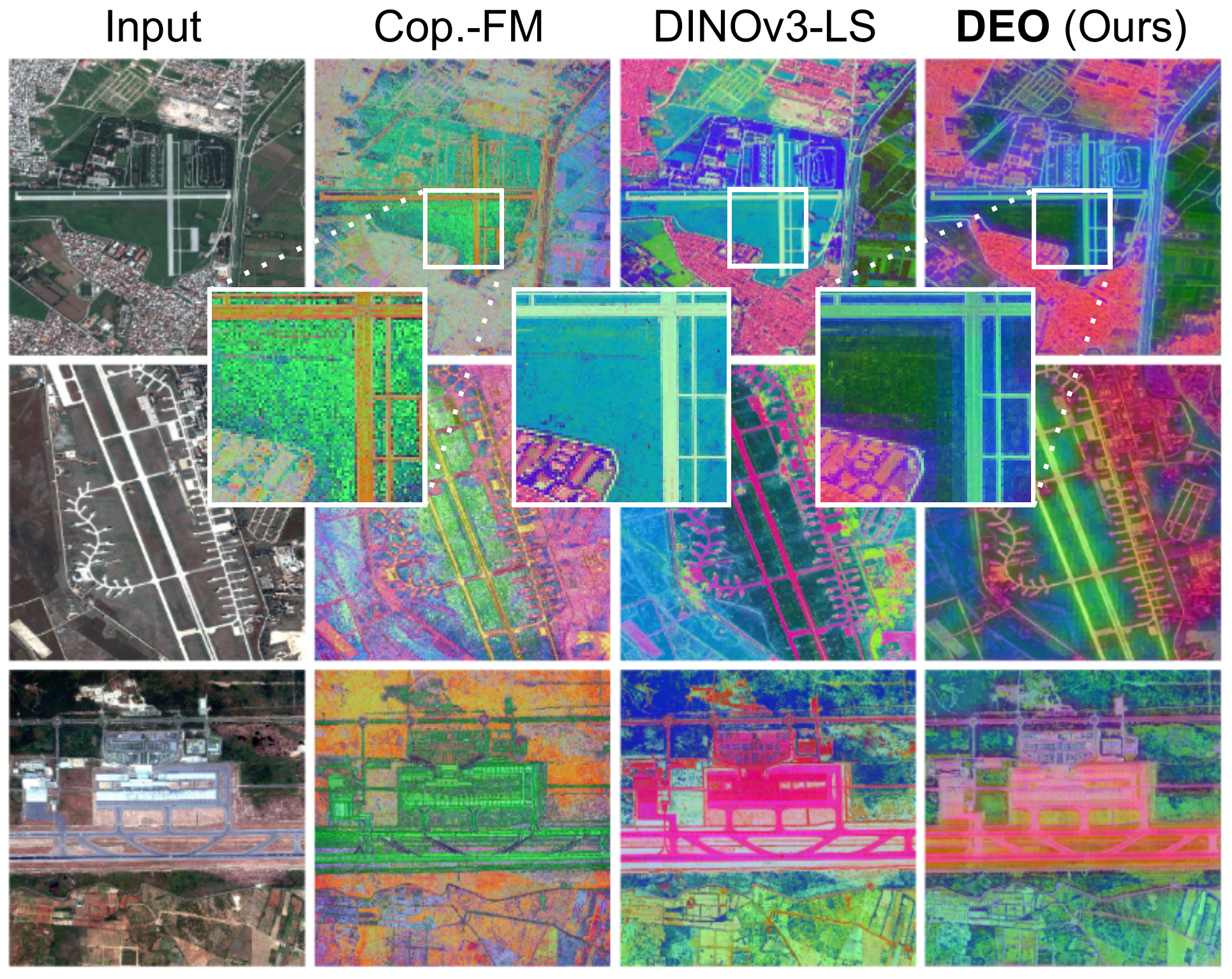}
   \caption{\textbf{PCA feature visualization and comparison} between Copernicus-FM~\cite{wang2025towards}, DINOv3-LS~\cite{simeoni2025dinov3}, and DEO (ours). We note the similarity of our method's features to those of DINOv3. We provide additional alignment experiments in the Supplementary.}
   \label{fig:pca}
   \vspace{-0.75em}
\end{figure}

\begin{table*}[t]
\centering
\resizebox{\textwidth}{!}{%
\setlength{\tabcolsep}{4pt}
\begin{tabular}{lccccccccccc}
\toprule
 & \multicolumn{5}{c}{\textcolor{optical}{Optical}} & \multicolumn{5}{c}{\textcolor{MS}{Multispectral}} &  \\
\cmidrule(lr){2-6} \cmidrule(lr){7-11}
Method & SN & GB-cattle & GB-pv & GB-chesa. & \textit{Avg} & GB-SA-c. & GB-cas. & S1F11 & PASTIS & \textit{Avg} & \textit{Overall Avg} \\
\midrule
DINOv2-B~\cite{oquab2024dinov}\tiny\textcolor{optical}{\textbf{RGB}} & \bm2{79.76} & 76.75 & 94.30 & 65.86 & 79.17 & 29.39 & 54.51 & 87.92 & 14.89 & 46.68 & 62.92 \\
DINOv3-B~\cite{simeoni2025dinov3}\tiny\textcolor{optical}{\textbf{RGB}} & 79.06 & 73.01 & 94.34 & 64.04 & 77.61 & 30.82 & 57.87 & 87.50 & 15.58 & 47.94 & 62.78 \\
DINOv3-LS~\cite{simeoni2025dinov3}\tiny\textcolor{optical}{\textbf{RGB}} & 78.67 & 72.86 & 94.04 & 60.03 & 76.40 & 33.05 & 63.32 & 88.78 & 16.45 & 50.40 & 63.40 \\
Scale-MAE~\cite{reed2023scale}\tiny\textcolor{optical}{\textbf{RGB}} & 76.24 & 51.66 & 91.45 & 33.33 & 63.17 & 23.13 & 59.42 & 84.77 & 6.70 & 43.50 & 53.33 \\
GFM~\cite{mendieta2023towards}\tiny\textcolor{optical}{\textbf{RGB}} & 77.30 & 62.12 & 93.56 & 70.49 & 75.87 & 29.79 & 62.59 & 85.42 & 18.66 & 49.12 & 62.49 \\
SatDiFuser~\cite{jia2025can}\tiny\textcolor{optical}{\textbf{RGB}} & 77.17 & \bm1{82.98} & \bm1{95.30} & \bm2{71.60} & \bm2{81.76} & 32.60 & \bm2{66.50} & 84.08 & 17.65 & 50.21 & \bm2{65.99} \\
CROMA~\cite{fuller2023croma}\tiny\textcolor{MS}{\textbf{MS}} & 71.82 & 74.37 & 88.00 & 54.43 & 72.16 & 34.27 & 59.91 & \bm2{94.16} & \bm2{24.66} & \bm2{53.25} & 62.70 \\
TerraFM~\cite{danish2025terrafm}\tiny\textcolor{MS}{\textbf{MS}} & 73.15 & 65.81 & 91.59 & 54.47 & 71.26 & 30.95 & 59.49 & 92.72 & 19.65 & 50.70 & 60.98 \\
Cop.-FM~\cite{wang2025towards}\tiny\textcolor{MS}{\textbf{MS}} & 75.45 & 68.88 & 93.56 & 55.81 & 73.43 & \bm2{34.67} & 55.71 & 92.58 & 21.49 & 51.11 & 62.27 \\
\textbf{DEO} (Ours)  & \bm1{80.83} & \bm2{80.21} & \bm2{94.41} & \bm1{72.47} & \bm1{81.98} & \bm1{37.96} & \bm1{68.56} & \bm1{94.32} & \bm1{28.96} & \bm1{57.45} & \bm1{69.72} \\
\bottomrule
\end{tabular}
}
\caption{\textbf{Results on \textcolor{optical}{optical} and \textcolor{MS}{multispectral} segmentation datasets}. Methods take as input either 3-channel \textcolor{optical}{optical} (\textcolor{optical}{RGB}) bands or multi-channel \textcolor{MS}{MS} bands. The GB prefix indicates that the dataset is from the GEO-Bench~\cite{lacoste2023geo} benchmark, SN represents SpaceNetv1~\cite{van2018spacenet}, S1F11 is Sen1Floods11~\cite{bonafilia2020sen1floods11}, and PASTIS is from \cite{garnot2021panoptic}. DINOv3-LS is the ViT-L version of DINOv3 pretrained on satellite imagery. All results are expressed in macro mIoU. \textcolor{goldD}{First} and \textcolor{silverD}{second} place results are marked. Further dataset details are available in the Supplementary.}
\label{tab:segmentation}
\vspace{-0.75em}
\end{table*}

\subsection{Incorporating optical knowledge}

The objective described in \Cref{subsec:ms} is well-suited for learning global MS representations, but falls short in producing the fine-grained, pixel-level features required for dense prediction tasks such as segmentation and change detection, as DINO-style reconstruction objectives emphasize global invariances without preserving local structure. Additionally, without dedicated optical supervision, the model does not develop specialized optical representations comparable to those learned by large optical VFMs~\cite{simeoni2025dinov3,waldmann2025panopticon}. To bridge this gap, we introduce a VFM as a second teacher, enabling the student to unify multispectral and optical knowledge within a single representation space. Unlike prior works that combine masked image modeling (MIM) with VFM distillation~\cite{mendieta2023towards, wang2025towards}, our objective from \Cref{eq:ms} is inherently compatible with the training paradigm of modern VFMs such as DINOv3, which also rely on contrastive self-distillation. As illustrated by the feature PCA visualization in \Cref{fig:pca} and additional experiments we provide in the Supplementary, the student's latent space aligns more closely to that of DINOv3 compared to the MIM-based Copernicus-FM, leading to more compatible feature transfer and improved downstream performance.

We introduce VFM distillation by expanding on the network described in \Cref{subsec:ms}. We pass $\textcolor{optical}{\text{I}^\text{O}_\text{g}}$ to the optical VFM teacher $\textcolor{optical}{\Phi_\text{O}}$ and $\textcolor{optical}{\text{I}^\text{O}_{\text{g}+\text{l}}}$ to the student $\textcolor{student}{\Phi_\text{s}}$, mirroring our MS learning part of the network whilst using a separate 3-channel patch embedding layer $\textcolor{optical}{pe^\text{O}}$ for optical data before the encoder. To incorporate both global and pixel-level knowledge from $\textcolor{optical}{\Phi_\text{O}}$, we distill the class token $\texttt{[cls]}_\text{F}$ and patch tokens $\texttt{[p]}_\text{F}$ from its final layer, as well as patch tokens $\texttt{[p]}_\text{mid}$ from an intermediate layer into corresponding student features. To achieve this, we introduce distinct projection heads $\textcolor{student}{p_\text{s}^\text{cls}}$, $\textcolor{student}{p_\text{s}^\text{p1}}$, and $\textcolor{student}{p_\text{s}^\text{p2}}$ for the student, which are separate from those used during MS training. This both aligns the feature dimension of the student with that of the teacher and decouples the projection heads between the optical and MS learning tasks, which has been shown to improve performance~\cite{oquab2024dinov}. We can then formulate the distillation loss for learning optical features as:
\begin{equation}
\begin{aligned}
\textcolor{optical}{\mathcal{L}_{O}} 
& = \alpha_1\mathcal{L}_\text{cos}((\textcolor{optical}{\boldsymbol{z}^\text{O}_\text{g}\texttt{[cls]}_\text{F}}),\,
      \textcolor{student}{p_\text{s}^\text{cls}}(\textcolor{student}{\boldsymbol{z}^\text{O}_{\text{g}+\text{l}}\texttt{[cls]}_\text{F}})) \\
& + \alpha_2\mathcal{L}_\text{cos}((\textcolor{optical}{\boldsymbol{z}^\text{O}_\text{g}\texttt{[p]}_\text{F}}),\,
      \textcolor{student}{p_\text{s}^\text{p1}}(\textcolor{student}{\boldsymbol{z}^\text{O}_{\text{g}+\text{l}}\texttt{[p]}_\text{F}})) \\
& + \alpha_3\mathcal{L}_\text{cos}((\textcolor{optical}{\boldsymbol{z}^\text{O}_\text{g}\texttt{[p]}_\text{mid}}),\,
      \textcolor{student}{p_\text{s}^\text{p2}}(\textcolor{student}{\boldsymbol{z}^\text{O}_{\text{g}+\text{l}}\texttt{[p]}_\text{mid}})),
\end{aligned}
\end{equation}
and we obtain the final learning loss by minimizing both the MS and optical learning objectives:
\begin{equation}
\mathcal{L} = -\textcolor{MS}{\mathcal{L}_{MS}} - \textcolor{optical}{\mathcal{L}_{O}}.
\end{equation}
We thereby train both multispectral and optical features in unison without compromising either feature space.

\subsection{Implementation details}

To aid in learning fine-grained features, we use the Swin~\cite{liu2021swin} transformer as our backbone, leveraging its hierarchical architecture and input patch size of $4$. Most VFMs use ViT~\cite{dosovitskiy2020image} with a patch size of $16$ as the standard backbone, limiting feature resolution. However, we demonstrate that distilling a ViT-based VFM into a Swin-based backbone yields fine-grained features that combine the knowledge of an optical teacher and a dedicated MS teacher.

For creating \textit{global} views of the input image, we crop in the range of $\{0.4, 1\}$ before resizing to an image size of $224\times224$. For \textit{local} views, we crop in the range of $\{0.05, 0.4\}$, and resize to $96\times96$. Following \cite{caron2021emerging}, we set $n=2$ and $m=10$. We set $\alpha_1=1$, $\alpha_2=0.5$, $\alpha_3=0.5$, and $\gamma=1$. Further training details are provided in the Supplementary.

\subsection{Pretraining}
\label{subsec:pretraining}

We pretrain our method for $100$ epochs using the Adam optimizer~\cite{kingma2014adam} and cosine learning rate scheduling on $16$ NVIDIA A100 GPUs, with a batch size of $8$ per GPU and a combined dataset of $500000$ images from the fMoW-Sentinel~\cite{cong2022satmae} and fMoW-RGB~\cite{christie2018functional} datasets. To construct the dataset, we first remove the three $60m$ resolution atmospheric bands from the 13-channel Sentinel-2 imagery in fMoW-Sentinel, leaving $10$ bands for pretraining. We then replace $150000$ low-spatial-resolution optical bands in Sentinel-2 imagery with their high-spatial-resolution aerial counterparts from fMoW-RGB at the same location, leading to improved performance on high-spatial-resolution datasets.
\section{Results}

We evaluate DEO on a diverse set of \textcolor{optical}{optical} and \textcolor{MS}{multispectral} segmentation, change detection, and classification datasets, comparing it against state-of-the-art methods. To more clearly differentiate methods, we specify whether they take as input multi-channel \textcolor{MS}{MS} data or only 3-channel \textcolor{optical}{optical} (RGB). Further evaluation details are available in the Supplementary.

\subsection{Segmentation}

To evaluate DEO's ability to employ fine-grained features, we report semantic segmentation results in \Cref{tab:segmentation} using UPerNet~\cite{xiao2018unified} as a segmentation head on a frozen backbone~\cite{jia2025can,tseng2025galileo,wang2025towards}. Methods are evaluated on the benchmark suite GEO-Bench~\cite{lacoste2023geo}, along with additional datasets SpaceNetv1~\cite{van2018spacenet}, Sen1Floods11~\cite{bonafilia2020sen1floods11}, and PASTIS~\cite{garnot2021panoptic}, creating a diverse mix of optical and MS datasets encompassing building, crop, flood, and animal segmentation. DEO achieves the best results in both modalities, with particularly remarkable average improvements of $4.20$ points over the state-of-the-art on MS datasets, thanks to our representation learning approach. This increase in MS performance is especially noticeable on tasks that benefit the most from MS data, such as crop and flood segmentation (GB-SA-c., PASTIS, and S1F11).

\begin{table}[t]
    \centering
    \begin{tabular}{lccc}
    \toprule
    ~ & \textcolor{optical}{Optical} & \textcolor{MS}{MS} & \\
    \cmidrule(lr){2-2} \cmidrule(lr){3-3}
      Method & LEVIR & OSCD & \textit{Avg}\\ 
      \midrule
      DINOv2-B~\cite{oquab2024dinov}\tiny\textcolor{optical}{\textbf{RGB}} & 91.1 & 49.0 & 70.1\\
      DINOv3-B~\cite{simeoni2025dinov3}\tiny\textcolor{optical}{\textbf{RGB}} & 91.6 & 55.2 & 73.4\\
      DINOv3-LS~\cite{simeoni2025dinov3}\tiny\textcolor{optical}{\textbf{RGB}} & \bm2{91.7} & 57.2 & 74.5\\
		 Scale-MAE~\cite{reed2023scale}\tiny\textcolor{optical}{\textbf{RGB}} & \bm1{92.1} & 47.2 & 69.6\\
      GFM~\cite{mendieta2023towards}\tiny\textcolor{optical}{\textbf{RGB}} & 89.8 & 54.1 & 72.0\\
      SatDiFuser~\cite{jia2025can}\tiny\textcolor{optical}{\textbf{RGB}} & 90.2 & 55.2 & 72.7\\
		 CROMA~\cite{fuller2023croma}\tiny\textcolor{MS}{\textbf{MS}} & 88.5 & 52.3 & 70.4\\
		 TerraFM~\cite{danish2025terrafm}\tiny\textcolor{MS}{\textbf{MS}} & 89.5 & 57.5 & 73.5\\
		 Cop.-FM~\cite{wang2025towards}\tiny\textcolor{MS}{\textbf{MS}} & 90.7 & \bm2{58.7} & \bm2{74.7}\\
		\textbf{DEO} (Ours)  & $91.3$ & \bm1{60.4} & \bm1{75.9}\\

    \bottomrule
    \end{tabular}
    \caption{\textbf{Results on \textcolor{optical}{optical} and \textcolor{MS}{multispectral} bi-temporal change detection datasets}. Methods take as input either 3-channel \textcolor{optical}{optical} (\textcolor{optical}{RGB}) bands or multi-channel \textcolor{MS}{MS} bands. All results are expressed in binary F1 score considering only the change class. \textcolor{goldD}{First} and \textcolor{silverD}{second} place results are marked.}
    \label{tab:cd}
    \vspace{-0.75em}
\end{table}

\subsection{Change detection}

We evaluate performance in remote sensing change detection using two well-established datasets: the optical LEVIR~\cite{chen2020levirStanet} and the multispectral OSCD~\cite{daudt2018urban} datasets. We follow the protocol from related work~\cite{rolih2025btc, wang2024mtp, mendieta2023towards} and extract backbone features from a pair of pre- and post-event images, then fuse them using element-wise subtraction. Fused features are processed using a UPerNet~\cite{xiao2018unified} decoder to produce the final binary change map. We report the results in \Cref{tab:cd} using the binary F1 metric considering change class only~\cite{rolih2025btc, wang2024mtp, mendieta2023towards}. DEO achieves the best performance in the MS setting, outperforming the previous best by $1.7$ points, setting a new state-of-the-art. It also achieves competitive performance on optical settings. While methods like Scale-MAE~\cite{reed2023scale} outperform it on optical data, DEO balances MS and optical performance more effectively, achieving the best overall results.

\begin{table}[t]
\centering
\resizebox{\linewidth}{!}{%
\begin{tabular}{lcccc}
\toprule
~ &  \multicolumn{4}{c}{\textcolor{MS}{Multispectral}} \\
\cmidrule(lr){2-5}
Method & GB-ben & GB-s2s & GB-es & \textit{Avg} \\
\midrule
DINOv2-B~\cite{oquab2024dinov}\tiny\textcolor{optical}{\textbf{RGB}} & 50.56 & 41.38 & 89.6 & 60.51 \\
DINOv3-B~\cite{simeoni2025dinov3}\tiny\textcolor{optical}{\textbf{RGB}} & 55.48 & \bm1{52.94} & 93.3 & \bm2{67.91} \\
DINOv3-LS~\cite{simeoni2025dinov3}\tiny\textcolor{optical}{\textbf{RGB}} & \bm2{58.70} & 48.68 & 92.4 & 66.59 \\
Scale-MAE~\cite{reed2023scale}\tiny\textcolor{optical}{\textbf{RGB}} & 46.06 & 44.02 & 92.4 & 60.83 \\
GFM~\cite{mendieta2023towards}\tiny\textcolor{optical}{\textbf{RGB}} & 51.91 & 47.26 & \bm2{95.5} & 64.89 \\
SatDiFuser~\cite{jia2025can}\tiny\textcolor{optical}{\textbf{RGB}} & 49.97 & 35.19 & 88.2 & 57.12 \\
CROMA~\cite{fuller2023croma}\tiny\textcolor{MS}{\textbf{MS}} & 53.13 & 46.65 & 89.3 & 63.69 \\
TerraFM~\cite{danish2025terrafm}\tiny\textcolor{MS}{\textbf{MS}} & \bm1{62.15} & 47.57 & 93.1 & 67.61\\
Cop.-FM~\cite{wang2025towards}\tiny\textcolor{MS}{\textbf{MS}} & 45.65 & 44.93 & 87.9 & 59.49 \\
\textbf{DEO} (Ours) & 58.43 & \bm2{52.23} & \bm1{97.0} & \bm1{69.22} \\
\bottomrule
\end{tabular}
}
\caption{\textbf{Results on \textcolor{MS}{multispectral} classification datasets with linear probing}. Methods take as input either 3-channel \textcolor{optical}{optical} (\textcolor{optical}{RGB}) bands or multi-channel \textcolor{MS}{MS} bands. The GB prefix indicates that the dataset is from the GEO-Bench~\cite{lacoste2023geo} benchmark. GB-ben (m-bigearthnet) results are expressed in the F1 metric, while GB-s2s (m-so2sat) and GB-es (m-eurosat) results are expressed in Top-1 accuracy. 
\textcolor{goldD}{First} and \textcolor{silverD}{second} place results are marked.}
\label{tab:classification}
\vspace{-0.75em}
\end{table}

\subsection{Classification}

To perform image classification, we extract class tokens or pooled patch tokens from the last layer of a frozen backbone and train a linear classifier on top of it. We report results on three diverse MS land cover datasets in \Cref{tab:classification}. Our model achieves the best overall results, with an average performance $1.3$ points above that of the next best-performing method. It also surpasses or closely matches the performance of the best-performing methods on individual datasets.

\begin{table}[t]
\centering
\resizebox{\linewidth}{!}{%
\setlength{\tabcolsep}{4pt}
\begin{tabular}{lccccc}
\toprule
& \multicolumn{3}{c}{Average Rank} & \multirow{2}{*}{\begin{tabular}[c]{@{}c@{}}Param.\\size (M)\end{tabular}} & \multirow{2}{*}{\begin{tabular}[c]{@{}c@{}}Pretrain\\size (M) \end{tabular}} \\ \cmidrule{2-4}
Method & \textcolor{optical}{Optical} & \textcolor{MS}{MS} & Overall &  &  \\
\midrule
Scale-MAE~\cite{reed2023scale}     & 8.0                                             & 8.3                                   & 8.2                        & 303                   & 0.36                \\
GFM~\cite{mendieta2023towards}     & 4.5                                             & 5.7                                   & 5.1                        & 87                   & 0.6                    \\
TerraFM~\cite{danish2025terrafm}   & 6.5                                             & 3.5                                   & 5.0                        & 85                    & 18                 \\
DINOv2-B~\cite{oquab2024dinov}    & 3.0                                             & 7.1                                   & 5.0                        & 85                    & 142                 \\
Cop.-FM~\cite{wang2025towards}     & 5.8                                             & 4.0                                   & 4.9                        & 139                   & 18                  \\
CROMA~\cite{fuller2023croma}       & 6.7                                             & \bm2{3.0}            & 4.8                        & 85                    & 1                         \\
DINOv3-B~\cite{simeoni2025dinov3}  & 3.4                                             & 6.0                                   & 4.7                        & 85                    & 1689                     \\
DINOv3-LS~\cite{simeoni2025dinov3} & 3.8                                             & 5.5                                   & 4.6                        & 303                   & 493                       \\
SatDiFuser~\cite{jia2025can}       & \bm2{2.0}                      & 4.7                                   & \bm2{3.3} & 949                    & 0.72                \\
\textbf{DEO} (ours)                     & \bm1{1.3}                      & \bm1{1.2}            & \bm1{1.3} & 87                   & 0.5 \\
\bottomrule
\end{tabular}
}
\caption{\textbf{Summary of methods and results.} Ranks for all tested methods over benchmark datasets, averaged over optical datasets, multispectral (MS) datasets, and overall. \textcolor{goldD}{First} and \textcolor{silverD}{second} place results are marked. Additionally, model size and pretraining corpus size are provided for reference.}
\label{tab:avg_ranks}
\vspace{-0.75em}
\end{table}

\begin{table*}[t]
\centering
\begin{minipage}[t]{0.65\textwidth}
\centering
\resizebox{\linewidth}{!}{%
\begin{tabular}{lcccccc}
\toprule
 & \multicolumn{2}{c}{\textcolor{optical}{Optical}} & \multicolumn{2}{c}{\textcolor{MS}{Multispectral}} & \multicolumn{2}{c}{Overall} \\
\cmidrule(lr){2-3} \cmidrule(lr){4-5} \cmidrule(lr){6-7}
Component & \textit{Avg} & Improv. & \textit{Avg} & Improv. & \textit{Avg} & Improv. \\
\midrule
Base (MS) & 77.87 &  & 60.44 &  & 69.16 &  \\
\hspace{0.5em}\small\textcolor{darkgray}{+ DINOv3\texttt{[cls]}}  & 79.07 & $\textcolor{darkgreen}{\uparrow}$ \textcolor{darkgreen}{1.20} & 62.81 & $\textcolor{darkgreen}{\uparrow}$ \textcolor{darkgreen}{ 2.37} & 70.94 & $\textcolor{darkgreen}{\uparrow}$ \textcolor{darkgreen}{1.79} \\
\hspace{0.5em}\small\textcolor{darkgray}{+ Sep. Opt. path} & 81.20 & $\textcolor{darkgreen}{\uparrow}$ \textcolor{darkgreen}{2.13} & 62.69 & $\textcolor{lightred}{\downarrow}$ \textcolor{lightred}{0.12} & 71.95 & $\textcolor{darkgreen}{\uparrow}$ \textcolor{darkgreen}{1.00} \\
\hspace{0.5em}\small\textcolor{darkgray}{+ DINOv3\texttt{[p]}} & 81.74 &$\textcolor{green}{\uparrow}$ \textcolor{green}{0.53} & 62.46 &$\textcolor{lightred}{\downarrow}$ \textcolor{lightred}{0.23} & 72.10 & $\textcolor{lightgreen}{\uparrow}$ \textcolor{lightgreen}{0.15} \\
\hspace{0.5em}\small\textcolor{darkgray}{+ Optical Aug.} & 81.95 &$\textcolor{lightgreen}{\uparrow}$ \textcolor{lightgreen}{0.22} & 63.02 &$\textcolor{green}{\uparrow}$ \textcolor{green}{0.55} & 72.48 & $\textcolor{lightgreen}{\uparrow}$ \textcolor{lightgreen}{0.39} \\
\hspace{0.5em}\small\textcolor{darkgray}{+ High Res. Optical} & \textbf{82.22} &$\textcolor{lightgreen}{\uparrow}$ \textcolor{lightgreen}{0.27} & \textbf{63.51} &$\textcolor{green}{\uparrow}$ \textcolor{green}{0.50} & \textbf{72.87} &$\textcolor{lightgreen}{\uparrow}$ \textcolor{lightgreen}{0.38} \\
\bottomrule
\end{tabular}
}
\caption{Ablation study of DEO components on three optical and three multispectral datasets. We present average results for each dataset category and the overall averages. We also show relative improvement for each component. Best results are marked in \textbf{bold}.}
\label{tab:ablation}
\end{minipage}
\hfill
\begin{minipage}[t]{0.32\textwidth}
\centering
\resizebox{\linewidth}{!}{%
\begin{tabular}{lccc}
\toprule
VFM & \textcolor{optical}{Optical} & \textcolor{MS}{MS} & Overall \\
\midrule
DINOv2~\cite{oquab2024dinov} & 85.01  & 62.71 & 73.86 \\
DINOv3~\cite{simeoni2025dinov3} & \textbf{85.03} & 62.77 & \textbf{73.90} \\
RADIOv2.5~\cite{heinrich2025radiov2} & 83.57 & \textbf{63.12} & 73.34 \\
\bottomrule
\end{tabular}
}
\caption{Comparison of distillation backbones. Best results for each dataset category and overall are presented in \textbf{bold}. Additional details are provided in the Supplementary.}
\label{tab:distillation}
\end{minipage}
\vspace{-0.75em}
\end{table*}

\subsection{Overall}

In \Cref{tab:avg_ranks}, we present the average rank for each tested method across all tasks on optical and MS datasets, as well as the overall average rank. DEO shows a strong overall lead, with a pronounced advantage in MS performance. The second-best-performing method on optical tasks, SatDiFuser~\cite{jia2025can}, significantly lags behind DEO in performance on MS tasks, despite having a model size more than $10$ times larger. Two other methods we tested utilize distillation: Copernicus-FM~\cite{wang2025towards} and GFM~\cite{mendieta2023towards}, with GFM also employing a Swin transformer. Due to our pretraining approach, described in \Cref{sec:method}, our method indirectly incorporates large amounts of data from VFMs in a manner that more carefully aligns feature spaces, resulting in a substantial performance lead over both methods, while using much less pretraining data. Remarkably, we outperform all tested versions of DINOv2~\cite{oquab2024dinov} and DINOv3~\cite{simeoni2025dinov3}, including the ViT-Large-based version pretrained on satellite data, demonstrating our method's ability to effectively incorporate optical VFM features with contrastive MS pretraining. In the Supplementary, we report the compute report and estimated pretraining emissions for DEO and DINOv3.

\subsection{Ablation study}

\noindent\textbf{Component ablation.} We introduce multiple components to our pretraining architecture to simultaneously learn \textcolor{MS}{MS} and \textcolor{optical}{optical} features without compromising either feature space. To better understand the impact of each component, we begin with a simple \textit{contrastive self-distillation baseline} designed for learning MS features, as presented in \Cref{subsec:ms} (Base (MS) in \Cref{tab:ablation}), and gradually build up to our state-of-the-art model. For consistency, each model is pretrained with a Swin-Tiny backbone on 50,000 images for 50 epochs and then fine-tuned on three optical (SN, GB-cattle, GB-pv) and three MS (GB-SA-c., GB-cas., S1F11) segmentation datasets. For each modality and overall, we present the average results, along with the relative improvements over the previous iteration, in \Cref{tab:ablation}. 

\begin{itemize}
    \item \textbf{DINOv3 distillation.} We begin by naively distilling the DINOv3\texttt{[cls]}~\cite{simeoni2025dinov3} token into the student features in addition to the contrastive multispectral objective. This results in a substantial increase in performance for both modalities, as we distill knowledge from a large VFM, despite aligning the MS feature space to resemble the optical representations derived from the VFM.
    
    \item \textbf{Separate optical path.} Introducing an additional projection head $\textcolor{student}{p_\text{s}^\text{cls}}$ and processing optical images in a separate pass with the student leads to a substantial improvement on optical tasks, while minimally degrading results for MS tasks. We attribute this to the optically derived DINOv3 features $\textcolor{optical}{\boldsymbol{z}^\text{O}_\text{g}\texttt{[cls]}_\text{F}}$ now being distilled into optically derived student features $\textcolor{student}{\boldsymbol{z}^\text{O}_{\text{g}\cup\text{l}}\texttt{[cls]}_\text{F}}$.
    
    \item \textbf{Patch token distillation.} To increase alignment with DINOv3 further, we additionally distill DINOv3\texttt{[patch]} tokens from the last and intermediate layers into last and intermediate stage student patch tokens $\textcolor{student}{p_\text{s}^\text{p1}}(\textcolor{student}{\boldsymbol{z}^\text{O}_{\text{g}\cup\text{l}}\texttt{[p]}_\text{F}})$ and $\textcolor{student}{p_\text{s}^\text{p2}}(\textcolor{student}{\boldsymbol{z}^\text{O}_{\text{g}\cup\text{l}}\texttt{[p]}_\text{mid}})$, leading to an improvement in optical performance and slight degradation in MS performance. While distilling the class token from DINOv3 incorporates general knowledge, distilling patch tokens helps our model mimic the out-of-the-box linearly separable patch features present in DINOv3~\cite{simeoni2025dinov3}.
    
    \item \textbf{Additional optical augmentations.} Introducing heavy augmentations $A_\text{O}$ for the optical input images  (\Cref{subsec:input}) leads to an overall improvement. While CL methods generally benefit from strong augmentations, this also helps incorporate \textit{robust} features from DINOv3, i.e., features that are less susceptible to input perturbations.
    
    \item \textbf{High-resolution optical data.} Replacing the low-resolution Sentinel-2 optical data with high-resolution aerial imagery leads to an overall increase in performance. This is especially noticeable on MS datasets, likely because high-resolution optical data is transferred to lower-resolution MS data, thereby serving as privileged knowledge. It also provides more detailed features, which are usually useful for dense tasks that require high resolution.
\end{itemize}
\noindent Our additions to the baseline training pipeline significantly increase performance by an average of $4.35$ points on optical and $3.07$ points on MS tasks.

\noindent\textbf{VFM distillation.} 
Results for various VFM optical teachers used for distillation are shown in \Cref{tab:distillation}, averaged over four optical and four MS datasets. Besides DINOv3~\cite{simeoni2025dinov3}, we evaluate DINOv2~\cite{oquab2024dinov} and RADIOv2.5~\cite{heinrich2025radiov2} which integrates features of multiple VFMs. DINOv2 and DINOv3 yield comparable results, although at large scales, distilling DINOv3 generally yields better performance. RADIOv2.5 performs slightly worse on optical datasets and better on MS datasets, but since the primary strength of VFMs lies in the optical domain, we opt to distill DINOv3 in our final model.

\begin{table}[t]
\centering
\resizebox{\linewidth}{!}{%
\begin{tabular}{lccc}
\toprule
& \multicolumn{3}{c}{10\%} \\
\cmidrule(lr){2-4}
Method & GB-SA-c. &   GB-BEN & SpaceNetv1 \\
\midrule

DINOv2-B~\cite{oquab2024dinov}\tiny\textcolor{optical}{\textbf{RGB}}& 23.04  & 44.43 & 75.57  \\
DINOv3-B~\cite{simeoni2025dinov3}\tiny\textcolor{optical}{\textbf{RGB}} & 25.22 & 46.52 & 74.36 \\
DINOv3-LS~\cite{simeoni2025dinov3}\tiny\textcolor{optical}{\textbf{RGB}} & 27.83 & 47.84 & \bm2{74.75} \\
Scale-MAE~\cite{reed2023scale}\tiny\textcolor{optical}{\textbf{RGB}} & 24.73 & 28.78 & 72.80 \\
GFM~\cite{mendieta2023towards}\tiny\textcolor{optical}{\textbf{RGB}} & 22.82 & 33.99 & 72.57 \\
SatDiFuser~\cite{jia2025can}\tiny\textcolor{optical}{\textbf{RGB}} &  20.87  & 43.75 &73.00\\
TerraFM~\cite{danish2025terrafm}\tiny\textcolor{MS}{\textbf{MS}} & 26.29 &  \bm2{48.53} & 69.24 \\
CROMA~\cite{fuller2023croma}\tiny\textcolor{MS}{\textbf{MS}} & 26.81 & 40.72 & 68.27 \\
Cop.-FM~\cite{wang2025towards}\tiny\textcolor{MS}{\textbf{MS}} & \bm2{28.15} & 28.71 & 71.65 \\
\textbf{DEO} (Ours) & \bm1{28.98} & \bm1{51.89} & \bm1{76.46} \\
\bottomrule
\end{tabular}
}
\caption{\textbf{Results in a low-data regime (10\%)}. Methods take as input either 3-channel \textcolor{optical}{optical} (\textcolor{optical}{RGB}) bands or multi-channel \textcolor{MS}{MS} bands. The GB prefix indicates that the dataset is from the GEO-Bench~\cite{lacoste2023geo} benchmark. GB-SA-c. and SpacenNetv1~\cite{van2018spacenet} results are expressed in macro mIoU, while GB-BEN results are expressed in F1. \textcolor{goldD}{First} and \textcolor{silverD}{second} place results are marked.}
\label{tab:low_data}
\vspace{-0.75em}
\end{table}

\noindent\textbf{Low-data regime.} We evaluate DEO under limited labeled data availability across three datasets. The results in \Cref{tab:low_data} demonstrate that our method maintains strong performance even with limited fine-tuning, validating our choice to utilize contrastive learning, as it generally requires less fine-tuning to perform well compared to other pretraining paradigms~\cite{shekhar2022understanding,balestriero2024learning}. This is especially practical for tasks where collecting labels is challenging, e.g., when rapid response is required during natural disasters.~\cite{meneses2022rapidMap, valsamis2024wildfire}. 

\subsection{Qualitative results}

\begin{figure}[h]
  \centering
   \includegraphics[width=\linewidth]{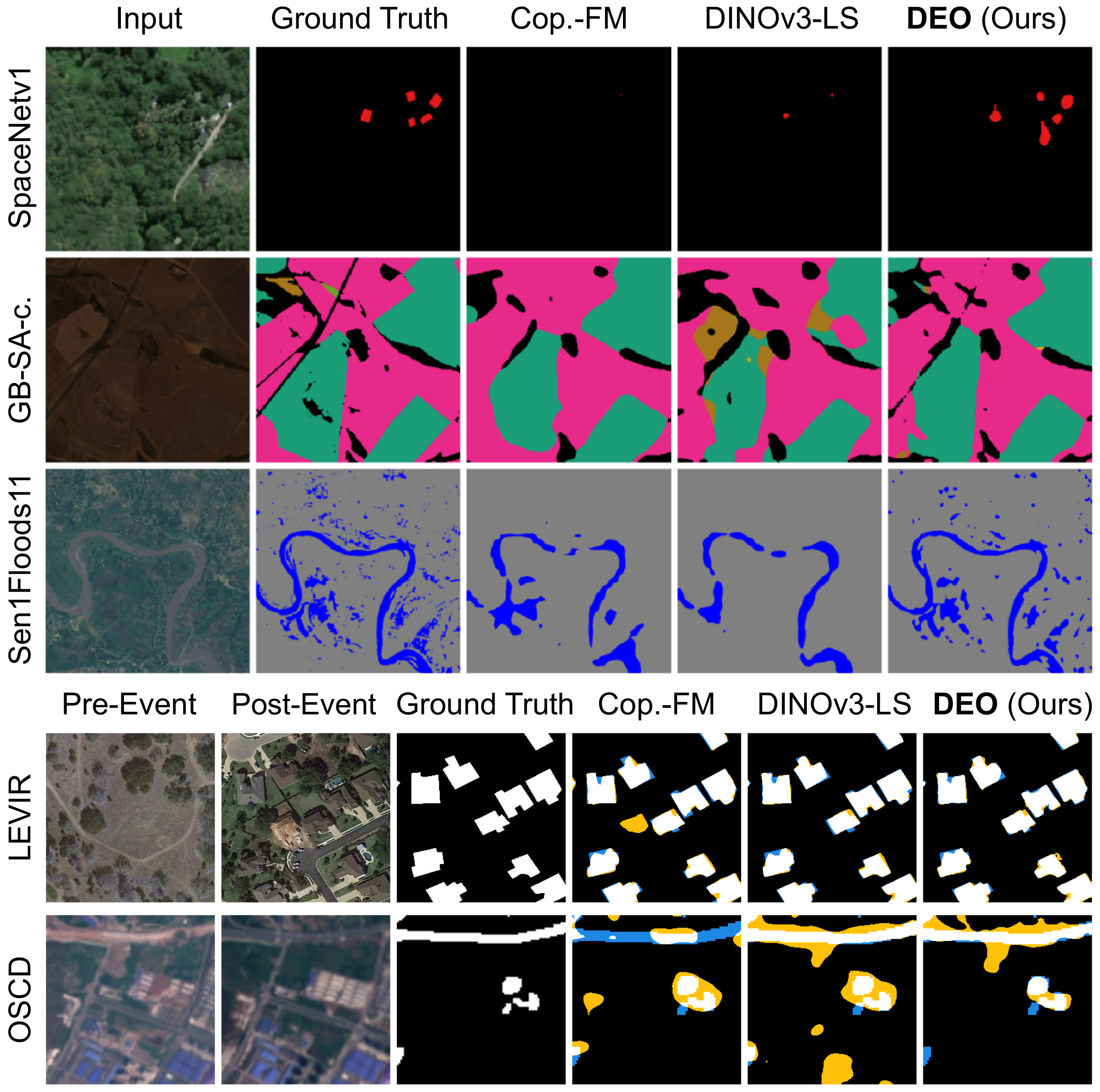}
   \caption{Qualitative results for semantic segmentation. The first two columns contain the optical part of the input image and the ground truth. The final three columns contain predictions from two related models, and our own.}
   \label{fig:seg}
   \vspace{-0.75em}
\end{figure}

Qualitative results for segmentation and change detection are shown in \Cref{fig:seg}. On SpaceNetv1~\cite{van2018spacenet} and Sen1Floods11~\cite{bonafilia2020sen1floods11}, DEO captures fine detail that competing methods fail to capture, while on GB-SA-c.~\cite{lacoste2023geo}, our segmentation mask is both finer in detail and more accurate. Performance on LEVIR~\cite{chen2020levirStanet} is comparable across models, but our method achieves higher precision on the multispectral OSCD dataset~\cite{daudt2018urban}. Unlike the optical-only DINOv3~\cite{simeoni2025dinov3}, DEO leverages information beyond the optical domain, reducing false positives. 
Compared to Copernicus-FM~\cite{wang2025towards}, it further improves utilization of multispectral information, demonstrating the effectiveness of our dual-teacher paradigm.

\noindent\textbf{Limitations.}
Our method is, to some degree, limited by the strength of the optical VFM used for distillation, as optical features are transferred rather than explicitly pretrained. It also assumes spatially aligned input, which is valid for optical and multispectral data, but poses a challenge for sensors from other platforms. Additionally, the lack of a strong teacher for modalities such as SAR limits the method's extensibility to such modalities. We plan to explore both sensor alignment and temporal alignment through pretraining tasks in the future.
\section{Conclusion}

We presented a dual-teacher distillation framework for multispectral Earth observation pretraining, DEO, combining a contrastive self-distillation teacher with a vision foundation model (VFM) teacher. This design efficiently transfers high-level semantic knowledge from an optical-only VFM while adapting it to multispectral data through a compatible contrastive learning objective. Unlike prior approaches that couple masked image modeling with VFM distillation, our formulation aligns the student’s training paradigm with that of modern VFMs such as DINOv3, which themselves rely on contrastive self-distillation to produce semantically structured feature spaces. This alignment leads to more coherent cross-modal feature transfer, resulting in improved performance on both optical and multispectral benchmarks, with an average improvement of $3.64$ points in semantic segmentation, $1.2$ points in change detection, and $1.31$ points in classification tasks. These findings highlight distillation-based pretraining as a scalable and resource-efficient path toward an interoperable ecosystem of Earth observation foundation models.

\section*{Acknowledgements}
This work was in part supported by the ARIS research projects J2-60045 (RoDEO) and GC-0006 (GeoAI), research programme P2-0214, and the supercomputing network SLING (ARNES, EuroHPC Vega). 


% \clearpage
% \setcounter{page}{1}
\maketitlesupplementary

\appendix

\section{Pretraining}

In this section, we will expand on the pretraining methodology and details introduced in Section 3. We will first provide more insight into our pretraining methodology, followed by details on hyperparameters to aid in reproducibility.

\subsection{Contrastive self-distillation}

The full loss for the \textit{coding rate regularizer}~\cite{wu2025simplifying} described in Section 3.1 can be formulated as:

\begin{equation}
\mathcal{L}_\text{CR} = -\frac{p + N B}{p N B}\,\frac{1}{V}\sum_{i=1}^{V}\log\det\!\left(\boldsymbol{I}_p + \frac{p}{B N \varepsilon}\, z_i^{\top} z_i\right),
\end{equation}
where $ z_i^{\top} z_i$ represents the covariance matrix, and $\log\det$ is calculated using the Cholesky expansion, i.e.,
\begin{equation}
\log\det(A) = 2 \sum_{i=1}^{p} \log L_{ii}.
\end{equation}
Here, $L_{ii}$ are diagonal elements of the matrix $L$ that satisfies the Cholesky decomposition $A = LL^{\top}$. $B$ represents the batch size, while $N$ is the number of used GPUs. $p$ is the dimension of the features $z$ after projection (256 for MS learning). Since $\mathcal{L}_\text{CR}$ is calculated only on the global student and teacher views, $V=2$. Finally, $\varepsilon$ is a small constant used for balancing, here $0.05$. The first factor $\frac{p + N B}{p N B}$ is a heuristic to balance the loss and can be adjusted accordingly.
    
Locality degrades in the contrastive-only baseline because DINO-style objectives emphasize global invariances without preserving local structure. Patch‑token reconstruction (as in DINOv3) could address this, but it lacks one of DEO's key benefits: distillation of semantic priors from a general VFM. In addition, DEO implicitly restores locality through patch‑token distillation, without the complexity of DINOv3’s reconstruction pipeline, thereby providing a more complete and efficient solution.

\begin{table}[h]
\centering
\begin{tabular}{lc}
\toprule
Parameter & Value  \\
\midrule
$n$ & 2 \\
$m$ & 10 \\
$\alpha_1$ & 1 \\
$\alpha_2$ & 0.5 \\
$\alpha_3$ & 0.5 \\
$\gamma$ & 1 \\
EMA & 0.996 \\
Cos scheduler WD & 0.04 \\
Base LR & 0.0005 \\
Warmup epochs & 10 \\
\bottomrule
\end{tabular}
\caption{Details of pretraining hyperparameters.}
\label{tab:hyperparams}
\end{table}

\begin{table}[h]
\centering
\begin{tabular}{lc}
\toprule
Element & Value  \\
\midrule
\midrule
Swin \\
\midrule
Input patch size & 4 \\
Embedding dim. & 128 \\
Windows size & 12 \\
\midrule
\midrule
Projection head \\
\midrule
Hidden dim. & 2048 \\
Bottleneck dim. (MS) & 256 \\
Bottleneck dim. (Optical) & 1024 \\
\bottomrule
\end{tabular}
\caption{Network element sizes.}
\label{tab:network_details}
\vspace{-1em}
\end{table}

\subsection{Hyperparameters}

In \Cref{tab:hyperparams} we provide details about the hyperparameters used during pretraining. In \Cref{tab:network_details} we provide details about the sizes of network elements.

\subsection{Distillation training data}

\textbf{DINOv2}~\cite{oquab2024dinov} introduces an automatic data curation pipeline to build a diverse dataset consisting of 142 million images. It emphasizes curation, deduplication, and filtering to remove near-duplicates, low‑quality content, and domain biases.\\
\noindent\textbf{DINOv3}~\cite{simeoni2025dinov3} builds upon the previous work of automatic data curation introduced in DINOv2, by introducing the larger LVD‑1689M dataset. It is a large, web‑curated dataset containing 1.689 billion images and providing a balanced representation of data available on the internet.\\
\noindent\textbf{RADIOv2.5}~\cite{heinrich2025radiov2} was not trained on any particular dataset, but is trained using agglomerative multi-teacher distillation. It distills from DINOv2~\cite{oquab2024dinov}, CLIP~\cite{radford2021learning}, and SAM~\cite{kirillov2023segment}, combining their features and thereby distilling the datasets they were originally trained on.

\begin{table*}[t]
\vspace{-1em}
\centering
\resizebox{\linewidth}{!}{%
\begin{tabular}{llcccccccccc}
\textbf{Segmentation} \\
\toprule
GEO-Bench~\cite{lacoste2023geo} & In paper & Image Size & \# Classes & Train & Val & Test & \# Bands & RGB res & Sensors \\
\midrule
m-pv4ger-seg & GB-pv  & 320$\times$320 & 2  & 3000 & 403  & 403  & 3  & 0.1 & RGB  \\
m-chesapeake-landcover & GB-chesa.  & 256$\times$256 & 7  & 3000 & 1000 & 1000 & 4  & 1.0 & RGBN  \\
m-cashew-plantation & GB-cas.    & 256$\times$256 & 7  & 1350 & 400  & 50   & 13 & 10.0 & Sentinel-2 \\
m-SA-crop-type & GB-SA-c.     & 256$\times$256 & 10 & 3000 & 1000 & 1000 & 13 & 10.0 & Sentinel-2   \\
m-nz-cattle  & GB-cattle     & 500$\times$500 & 2  & 524  & 66   & 65   & 3  & 0.1 & RGB   \\
\midrule
Others \\
\midrule
SpaceNetv1~\cite{van2018spacenet} &  SN   & 224$\times$224 & 2  & 5000 & 1000  & 1000  & 3  & 0.5 & DigitalGlobe WorldView 2  \\
Sen1Floods11~\cite{bonafilia2020sen1floods11} & S1F11  & 512$\times$512 & 3  & 252 & 89 & 90 & 13  & 10.0 & Sentinel-2  \\
PASTIS~\cite{garnot2021panoptic} & PASTIS   & 128$\times$128 & 20  & 1455 & 482  & 496   & 10 & 10.0 & Sentinel-2 \\
\bottomrule
\end{tabular}
}
\caption{Details for segmentation datasets used in the paper for evaluation.}
\label{tab:seg_details}
\vspace{-1em}
\end{table*}

\begin{table*}[t]
\centering
\resizebox{\linewidth}{!}{%
\begin{tabular}{llcccccccccc}
\textbf{Classification} \\
\toprule
GEO-Bench~\cite{lacoste2023geo} & In paper & Image Size & \# Classes & Train & Val & Test & \# Bands & RGB res & Sensors \\
\midrule
m-bigearthnet &  GB-ben   & 120$\times$120 & 43  & 20000 & 1000  & 1000  & 12  & 10.0 & Sentinel-2  \\
m-so2sat & GB-s2s  & 32$\times$32 & 17  & 19992 & 986 & 986 & 18  & 1.0 & Sen.-2 + Sen.-1  \\
m-eurosat & GB-es    & 64$\times$64 & 10  & 2000 & 1000  & 1000   & 13 & 10.0 & Sentinel-2 \\
\bottomrule
\end{tabular}
}
\caption{Details for classification datasets used in the paper for evaluation.}
\label{tab:cls_details}
\vspace{-1em}
\end{table*}

\begin{table*}[!t]
\centering
\resizebox{\linewidth}{!}{%
\begin{tabular}{llcccccccccc}
\textbf{Change detection} \\
\toprule
GEO-Bench~\cite{lacoste2023geo} & In paper & Image Size & \# Classes & Train & Val & Test & \# Bands & RGB res & Sensors \\
\midrule
LEVIR-CD~\cite{chen2020levirStanet} &  LEVIR  & 256$\times$256 & 2  & 7120 & 1024  & 2048  & 3  & 0.5 & Google Earth satellite  \\
OSCD~\cite{daudt2018urban} & OSCD  & 96$\times$96 & 2  & 827 & - & 385 & 10  & 10.0 & Sentinel 2  \\
\bottomrule
\end{tabular}
}
\caption{Details for change detection datasets used in the paper for evaluation.}
\label{tab:cd_details}
\vspace{-1em}
\end{table*} 

\section{Evaluation}

In this section, we provide additional details on the datasets used for evaluation, along with a more detailed description of our evaluation methodology.

\subsection{Datasets}

\textbf{Semantic segmentation.} For semantic segmentation, we use a mixture of established benchmarks in the form of GEO-Bench~\cite{lacoste2023geo} and standalone datasets. Details are provided in \Cref{tab:seg_details}.\\
\noindent\textbf{Classification.} We use three multispectral classification datasets from the GEO-Bench~\cite{lacoste2023geo} benchmark. m-bigearthnet is a multi-label classification task, while m-so2sat and m-eurosat are single-label classification tasks. We provide details in \Cref{tab:cls_details}.\\
\noindent\textbf{Change detection.} We use the optical LEVIR~\cite{chen2020levirStanet} and multispectral OCSD~\cite{daudt2018urban} change detection datasets. Details are provided in \Cref{tab:cd_details}.

\subsection{Evaluation details}

For all experiments, we use a batch size of 64 and fine-tune for 50 epochs using a learning rate. We provide other details in the following section.

\noindent\textbf{Semantic segmentation.} For all methods, we fine-tune an UPerNet~\cite{xiao2018unified} segmentation head on top of a frozen backbone. We extract features from four stages of the backbone: for ViT-B backbones, these are stages 3, 5, 8, and 11; for ViT-L, stages 7, 11, 15, and 23; and for Swin-based backbones, we take the four Swin stages before pooling. For ViT backbones, the latest stage is downsampled, while the first and second stages are upsampled four times and 2 times, respectively. UPerNet details are provided in \Cref{tab:upernet}.\\
\noindent\textbf{Classification.} We extract the last layer features from a frozen backbone and train a simple linear layer on top of them. For ViT backbones, we extract the last-layer class token; for Swin backbones, we pool the last-layer features to simulate a class token.\\
\noindent\textbf{Change detection.}
For all evaluated methods, we first perform backbone feature fusion of a pair of images using a simple element-wise subtraction. We then pass them to a UPerNet~\cite{xiao2018unified}, except for ViT-based methods, where we use the UNet~\cite{ronneberger2015u} decoder, as it performs better. Following related work~\cite{wang2024mtp, rolih2025btc}, we also train the backbone for change detection. We extract features from four stages for all models: for ViT-B backbones, stages 3, 5, 7, and 11; for ViT-L, stages 7, 11, 15, and 23; and for Swin-based backbones, the four Swin stages before pooling. UPerNet details are provided in \Cref{tab:upernet}, and for UNet, we use the same setup as in~\cite{rolih2025btc}.

\begin{table}[h]
\vspace{-1em}
\centering
\begin{tabular}{lc}
\toprule
Parameter & Value  \\
\midrule
Pool scales & 1, 2, 3, 6 \\
Hidden size & 512 \\
\bottomrule
\end{tabular}
\caption{Details of the UPerNet segmentation head.}
\label{tab:upernet}
\vspace{-1em}
\end{table}

\section{Method details}

We implement each evaluated method using its official repository. We use consistent learning rates per method, except for SatDiFuser~\cite{jia2025can}, for which we use the provided learning rates. Learning rates are presented in \Cref{tab:lr}. Official repositories of evaluated methods are listed in the following:

\begin{itemize}
    \item DINOv2~\cite{oquab2024dinov}: https://github.com/facebookresearch/dinov2
    \item DINOv3~\cite{simeoni2025dinov3}: https://github.com/facebookresearch/dinov3
    \item Scale-MAE~\cite{reed2023scale}: https://github.com/bair-climate-initiative/scale-mae
    \item GFM~\cite{mendieta2023towards}: https://github.com/mmendiet/GFM
    \item SatDiFuser~\cite{jia2025can}: https://github.com/yurujaja/SatDiFuser
    \item CROMA~\cite{fuller2023croma}: https://github.com/antofuller/CROMA/tree/main
    \item TerraFM~\cite{danish2025terrafm}: https://github.com/mbzuai-oryx/TerraFM/blob/master/terrafm.py
    \item Copernicus-FM~\cite{wang2025towards}: https://github.com/zhu-xlab/Copernicus-FM
\end{itemize}

\begin{table}[h]
\vspace{-1em}
\centering
\begin{tabular}{lcc}
\toprule
& \multicolumn{2}{c}{LR} \\
\cmidrule(lr){2-3}
Dataset & Other methods &  SatDiFuser \\
\midrule
m-pv4ger-seg & $10^{-4}$ & $10^{-2}$\\
m-chesapeake-landcover & $10^{-4}$ & $10^{-2}$\\
m-cashew-plantation  & $10^{-2}$ & $10^{-2}$\\
m-SA-crop-type  & $10^{-4}$ & $10^{-2}$\\
m-nz-cattle & $10^{-4}$ & $10^{-3}$\\
SpaceNetv1  & $10^{-4}$ & $10^{-2}$\\
Sen1Floods11  & $10^{-4}$ & $10^{-2}$\\
PASTIS & $10^{-1}$ & $10^{-2}$\\
m-bigearthnet & $10^{-3}$ & $10^{-2}$\\
m-so2sat & $10^{-3}$ & $10^{-4}$\\
m-eurosat  & $10^{-2}$ & $10^{-2}$\\
LEVIR-CD  &  $10^{-4}$ & $10^{-4}$\\
OSCD   & $10^{-4}$ & $10^{-4}$\\
\bottomrule
\end{tabular}
\caption{Learning rates for datasets and methods.}
\label{tab:lr}
\vspace{-1em}
\end{table}

\section{Additional experiments}

In this section, we provide additional experiments to support and ablate various claims from the paper.

\subsection{Latent space alignment}

We conduct a quantitative Central Kernel Alignment analysis to provide additional support for the superior latent-space alignment of our method with DINOv3 compared to MIM-based methods. On optical inputs, DEO aligns more closely with DINOv3 than Copernicus-FM (0.65 vs. 0.49), supporting our claim that objective-level compatibility with contrastive self-distillation enables effective optical knowledge transfer. On MS inputs, alignment with DINOv3 is lower (0.15 vs. 0.50), \textit{reflecting the intended integration of additional spectral information beyond RGB}. This behavior is consistent with the use of the CR loss, which promotes integration of MS information and improves MS performance. Replacing DEO's contrastive objective with MIM (all else fixed) reduces both alignment and downstream performance (\Cref{tab:additional_experiments}).
% The visual artifacts in Fig. 3 are not upsampling effects but stem from Swin’s shifted-window attention, which can introduce minor patch-boundary inconsistencies. We will clarify these points in the revision.

\subsection{Coding rate experiments}

We perform experiments without the CR loss (\Cref{tab:additional_experiments}). The loss is applied to the MS branch; therefore, its omission significantly reduces MS performance. Total collapse is still avoided because the VFM teacher serves as an additional regularizer.

\subsection{Coding rate experiments}

The pretraining experiments with artificial shift (\Cref{tab:additional_experiments}) show that our model is robust to misalignment.

\subsection{ViT model}

We train a ViT-S (patch size 8) with a parameter count comparable to Swin-T. Its performance is slightly lower than the Swin model (\Cref{tab:additional_experiments}), mainly due to a few challenging datasets (GB-chesa., GB-cas.).

\begin{table}[h]
\vspace{-1em}
\centering
\begin{tabular}{llccc}
\toprule
Size & Model & \textcolor{optical}{Optical} & \textcolor{MS}{MS} & Overall \\
\midrule
 \multirow{3}{*}{\makecell{\textit{Large}\\\textit{models}}} & DEO (contrast.) & 81.98 & 57.45 & 69.72 \\
                                                              & DEO (MIM)       & 78.18 & 56.39 & 67.28 \\
                                                            & DEO (no CR)     & 82.02 & 54.47 & 68.25 \\
\midrule
 \multirow{3}{*}{\makecell{\textit{Small}\\\textit{models}}} & DEO (Swin-T)    & 79.88 & 55.45 & 67.67 \\
                                                             & DEO (ViT-S p8)  & 76.62 & 52.09 & 64.35 \\
                                              & DEO (shift)     & 79.86 & 55.56 & 67.71 \\
\bottomrule
\end{tabular}
\caption{Additional experiments.}
\label{tab:additional_experiments}
\vspace{-1em}
\end{table}

\section{Compute and carbon footprint}

We provide the compute report and estimated pretraining emissions using \textit{carbontracker} in \Cref{tab:compute}. DEO's inference performance is comparable to DINOv3-B, which has a similar parameter count. Patch size 4 in our Swin-based DEO yields only a small increase in compute due to windowed attention, whereas a ViT with a reduced patch size shows a larger increase.

\begin{table}[!h]
\vspace{-1em}
\centering
\resizebox{\linewidth}{!}{%
% \setlength{\tabcolsep}{4pt}
% \renewcommand{\arraystretch}{1.2}
% \begin{footnotesize}
\begin{tabular}{lcccc}
\toprule
 Model & \makecell[c]{img/s} & \makecell[c]{Mem. [MB]} & TFLOPS & \makecell[c]{kgCO2eq} \\
\midrule
% DEO (Swin-B) & 169 & 5272 & 10475 & 34.5 \\
% DEO (ViT-S p8) & 411 & 18152 & 31302 & x \\
% DINOv3-B & 102  & 5595 & 9823 & 18000 \\
 DEO (Swin-\textbf{B}) & 5.917 & 5272 & 10475 & 34.5 \\
 DINOv3-\textbf{B} & 9.80  & 5595 & 9823 & 18000 \\
\midrule
DEO (ViT-\textbf{S} p8) & 2.43 & 18152 & 31302  \\
\bottomrule
\end{tabular}
% \end{footnotesize}
}
\caption{Inference experiments.}
\label{tab:compute}
\vspace{-1em}
\end{table}

\section{Additional qualitative}

\begin{figure}[ht]
  \centering
   \includegraphics[width=\linewidth]{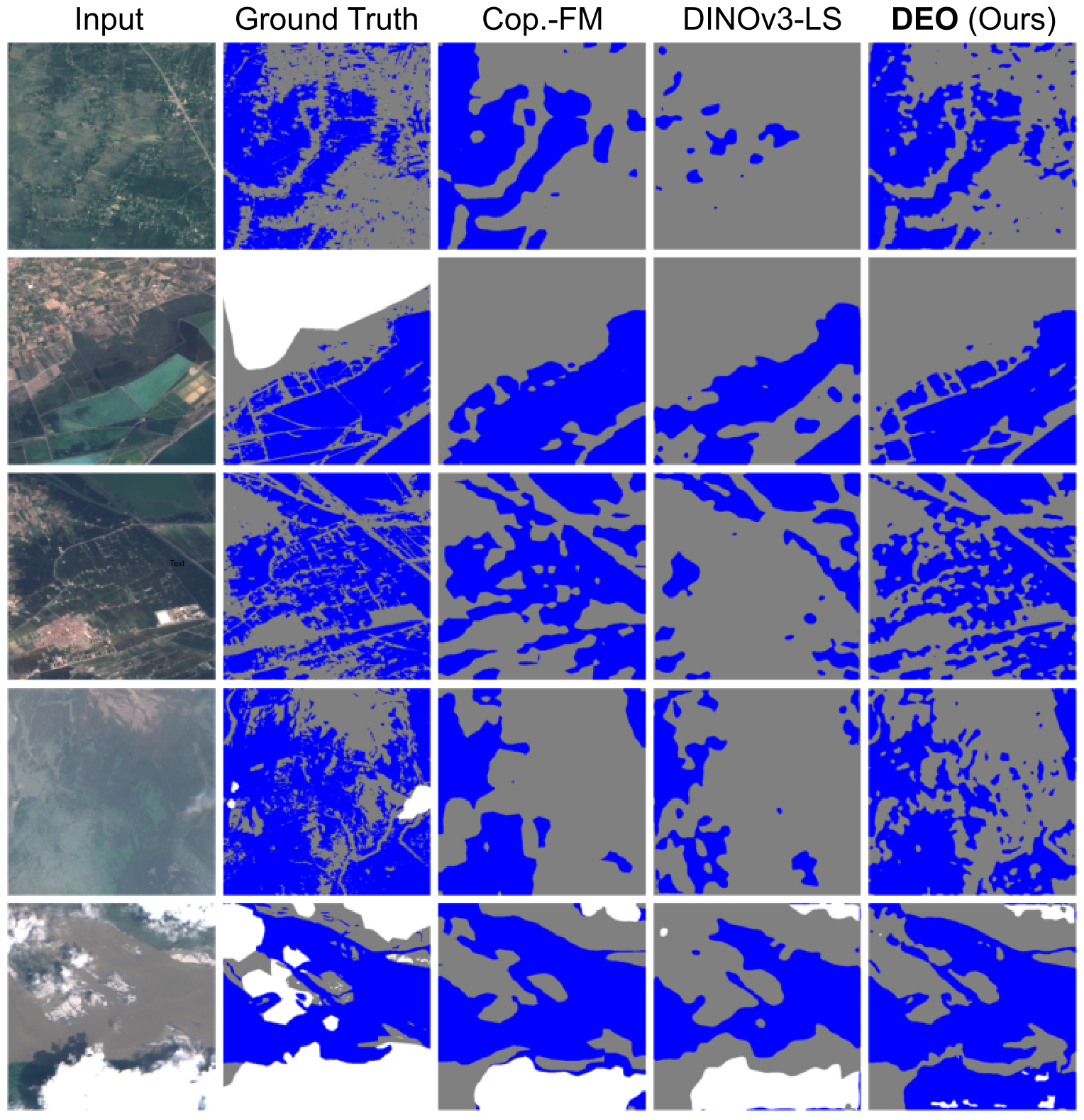}
   \caption{Extended qualitative results for Sen1Floods11.}
   \label{fig:qual_floods}
   % \vspace{-1em}
\end{figure}

\begin{figure}[ht]
% \vspace{-1em}
  \centering
   \includegraphics[width=\linewidth]{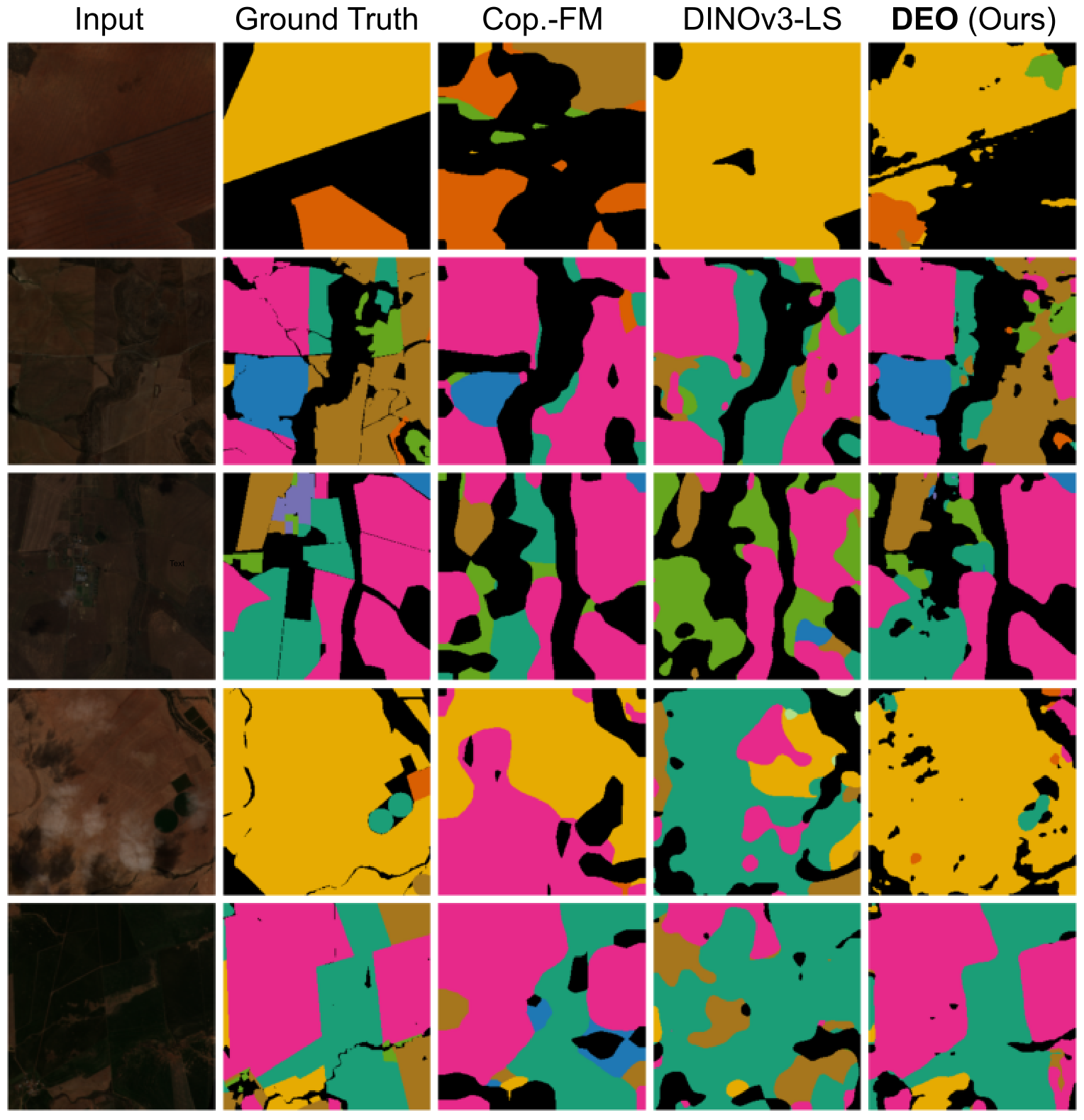}
   \caption{Extended qualitative results for m-SA-crop-type.}
   \label{fig:qual_crop}
   % \vspace{-1em}
\end{figure}

\begin{figure}[ht]
% \vspace{-1em}
  \centering
   \includegraphics[width=\linewidth]{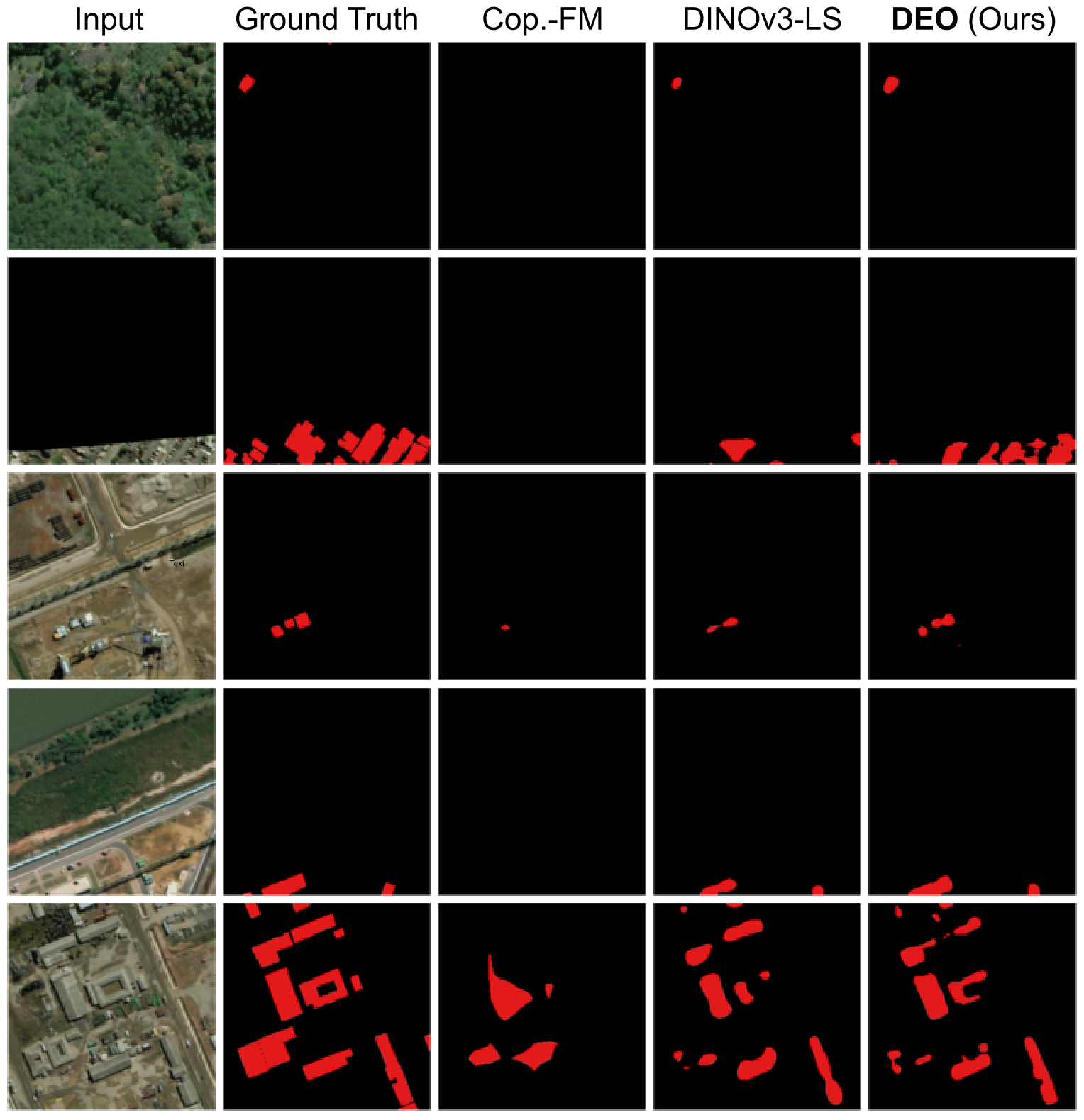}
   \caption{Extended qualitative results for SpaceNetv1.}
   \label{fig:qual_spacenet}
   % \vspace{-1em}
\end{figure}

\begin{figure}[ht]
% \vspace{-1em}
  \centering
   \includegraphics[width=\linewidth]{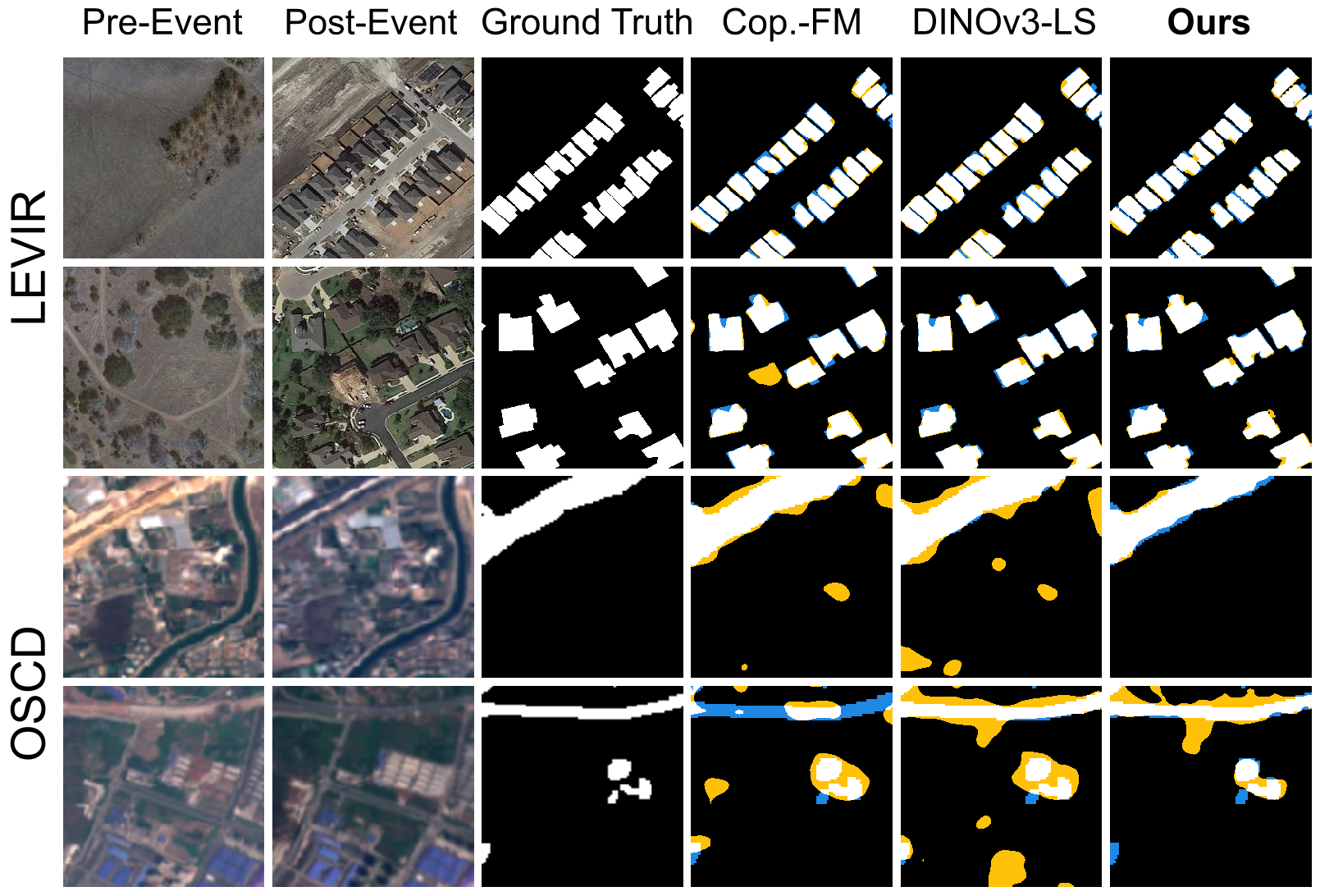}
   \caption{Extended qualitative results for LEVIR and OSCD.}
   \label{fig:qual_cd}
   % \vspace{-1em}
\end{figure}

We provide additional qualitative results in \Cref{fig:qual_floods,fig:qual_crop,fig:qual_spacenet,fig:qual_cd}.

\bibliographystyle{ieeenat_fullname}
{\small
\bibliography{main}
}
\end{document}